\documentclass[anon]{l4dc2027}

\usepackage{macros}
\usepackage{times}
\usepackage{tikz}
\usepackage{cleveref}
\usepackage{booktabs,array,float,placeins}
\usepackage[size=small]{caption}
\usetikzlibrary{arrows.meta,calc,positioning,decorations.pathreplacing}

\title[Feedback Linearised Robust Geometric DDPC]{Feedback-Linearised Trajectory Tracking via\\Model-Informed Robust Geometric DDPC}

\author{
    \Name{Pranav Gupta} \Email{guptapranav@iitb.ac.in}\\
    \addr Indian Institute of Technology Bombay
    \AND
    \Name{Deepan Muthirayan} \Email{deepan.muthirayan@plaksha.edu.in}\\
    \addr Plaksha University
    \AND
    \Name{Ravi Banavar} \Email{banavar@iitb.ac.in}\\
    \addr Indian Institute of Technology Bombay
}

\begin{document}

\maketitle

\begin{abstract}
    Data-driven predictive control (DDPC) predicts system responses from recorded trajectories, but nonlinear dynamics can limit the accuracy of fixed linear predictors. This paper studies geometric robust DDPC for trajectory tracking of a fully actuated manipulator using model-based feedback linearisation. A mechanical model maps virtual acceleration commands into joint torques, while recorded data capture residual dynamics arising from imperfect model parameters. The controller minimises worst-case prediction error over an open Grassmannian ball centred at the fitted trajectory subspace. Simulations on a planar 2R manipulator under gravity compare the proposed method with double-integrator model predictive control (MPC) and direct regularised data-enabled predictive control (DeePC). Results show improved tracking over double-integrator MPC despite increased torque effort, while performance relative to DeePC depends on the reference trajectory.
\end{abstract}

\begin{keywords}
    data-driven control, feedback linearisation, robust control, trajectory tracking.
\end{keywords}

\section{Introduction}
\label{sec:introduction}

Predictive control chooses future inputs by predicting the system's response, applies the first input, and repeats the optimisation after each new measurement. Model predictive control (MPC) predicts with a mathematical model. Data-driven predictive control (DDPC) predicts from recorded data instead, as in data-enabled predictive control (DeePC), which represents trajectories as linear combinations of recorded input--output data~\citep{coulson2019data}. The Fundamental Lemma justifies this representation for controllable linear time-invariant (LTI) systems with noiseless, sufficiently exciting data~\citep{willems2005note}.

Nonlinear systems with uncertain parameters challenge both: inaccurate mechanical models yield biased predictions, while linearly combining measured trajectories approximates nonlinear dynamics only locally. Model knowledge reduces the data burden. For fully actuated manipulators, feedback linearisation maps uses a mechanical model to map virtual acceleration commands to joint torques. With exact parameters and states, each joint becomes a double integrator from virtual input to position. Parameter, state-estimation and actuator errors produce residual dynamics learnable from virtual-input and position recordings.

Geometric robust DDPC~\citep{bharadwaj2026robust} fits a low-dimensional trajectory subspace and handles fitting errors using an open chordal-distance ball of same-dimensional subspaces on the Grassmannian. It minimises worst-case weighted squared deviation from a target combining measured history and desired future trajectory. The mechanical model supplies the feedback-linearising torque map; DDPC predicts the virtual-input-to-position response. We test whether learning this response improves prediction and tracking beyond the known double-integrator model.

\paragraph{Related work.}
Regularised and robust formulations extend DeePC to noisy data~\citep{coulson2019regularized,huang2023robust}. For nonlinear systems, \citet{berberich2022linear2} obtain closed-loop stability guarantees by updating the data of a linear predictor online. For feedback-linearisable systems, \citet{alsalti2023data} build a data-based predictor from input--output data and basis functions that approximate the unknown nonlinearities, with a fully actuated double inverted pendulum as example. \citet{watson2025hybrid} incorporates partial model knowledge into DeePC by combining model-based and data-based representations. Tube MPC handles bounded disturbances around a model prediction~\citep{mayne2005robust}, and learning-based MPC augments a model with learned corrections and uncertainty estimates~\citep{hewing2020learning,hewing2020cautious}.

\paragraph{Contributions.}
We apply geometric robust DDPC to trajectory tracking of a planar 2R manipulator under gravity. Prediction tests compare the learned feedback-linearised response with the ideal double integrator, for exact and inaccurate mechanical parameters. Transfer tests examine whether fixed parameter tunings remain effective when the measurement noise and arm parameters change. Tracking tests compare accuracy and torque effort with double-integrator MPC and with direct regularised DeePC, which predict from recorded joint torques and positions directly.

\section{Preliminaries}
\label{sec:preliminaries}


\paragraph{Notation.}
Operator \(\col(\cdot)\) stacks its arguments vertically. For \(Q\succeq0\) we write \(\norm{z}_Q^2=z^\top Qz\). We use \(\norm{\cdot}_{\mrm F}\) for the Frobenius norm, \(A^\dagger\) for the Moore--Penrose pseudoinverse, and \(\im A\) for the column space. \(I_r\) and \(\mbf1_r\) are the identity matrix and the all-ones vector, and \(\otimes\) is the Kronecker product.
For \(z_j\in\R[n_z]\), we write \(\mbf z_{a:b}=\col(z_a,\ldots,z_b)\). Superscripts \(\mrm d\), \(\mrm{ini}\) and \(\mrm{ref}\) mark recorded data, initial histories and references. The measured output is \(\wt y_k=y_k+\varepsilon_k^y\), where \(\varepsilon_k^y\) is measurement noise.
A record \(\mbf z^{\mrm d}=\mbf z^{\mrm d}_{0:T_{\mrm d}-1}\) of length \(T_{\mrm d}\) defines, for \(1\leq s\leq T_{\mrm d}\), the depth-\(s\) block Hankel matrix
\begin{equation}
    \mathcal H_s(\mbf z^{\mrm d})=
    \begin{bmatrix}
        z_0^{\mrm d}     & z_1^{\mrm d} & \cdots & z_{T_{\mrm d}-s}^{\mrm d}   \\
        z_1^{\mrm d}     & z_2^{\mrm d} & \cdots & z_{T_{\mrm d}-s+1}^{\mrm d} \\
        \vdots           & \vdots       & \ddots & \vdots                      \\
        z_{s-1}^{\mrm d} & z_s^{\mrm d} & \cdots & z_{T_{\mrm d}-1}^{\mrm d}
    \end{bmatrix}
    \in\R[n_zs\times(T_{\mrm d}-s+1)].
    \label{eq:preliminaries:hankel}
\end{equation}
Each column is a window of \(s\) consecutive samples. The record is \emph{persistently exciting of order \(s\)} if this matrix has full row rank, that is, \(\rank\mathcal H_s(\mbf z^{\mrm d})=n_zs\).

\subsection{Linear Systems and Model Predictive Control}
\label{sec:preliminaries:mpc}

Consider a controllable and observable discrete-time LTI system \(x_{k+1}=Ax_k+Bu_k\), \(y_k=Cx_k+Du_k\), with \(x_k\in\R[n_x]\), \(u_k\in\R[n_u]\) and \(y_k\in\R[n_y]\). Its lag \(\ell\) is the smallest positive integer for which the observability matrix \(\mathcal O_\ell=\col(C,CA,\ldots,CA^{\ell-1})\) has rank \(n_x\).

MPC predicts over a horizon of \(N\) samples. At step \(k\), let \(u_{k+i\mid k}\) and \(y_{k+i\mid k}\) denote the \((k+i)^{th}\) prediction, stacked as \(\mbf u_k\) and \(\mbf y_k\) over \(i=0,\ldots,N-1\). With desired sequences \(u_j^{\mrm{ref}},y_j^{\mrm{ref}}\), the input-output tracking cost is writen as
\begin{equation}
    J_k(\mbf u_k,\mbf y_k)=\sum_{i=0}^{N-1}\left(
    \norm{y_{k+i\mid k}-y_{k+i}^{\mrm{ref}}}_Q^2+
    \norm{u_{k+i\mid k}-u_{k+i}^{\mrm{ref}}}_R^2\right),
    \label{eq:preliminaries:tracking-cost}
\end{equation}
where \(Q\succeq0\) and \(R\succ0\). MPC minimises this cost subject to the dynamics and an initial state estimate, then applies the first input.

\subsection{Data-Based Prediction}
\label{sec:preliminaries:behaviour}

For \(w_j=\col(u_j,y_j)\), the length-\(L\) behaviour \(\mathcal B_L\) is the set of trajectories \(\mbf w=\col(w_0,\ldots,w_{L-1})\) that the system generates from arbitrary initial states and inputs. For \(L\geq\ell\), it is a subspace of dimension \(n_uL+n_x\)~\citep{markovsky2021behavioral}. Let \(\mbf w^{\mrm d}\) be a noiseless trajectory of the system with persistently exciting input of order \(L+n_x\). \citet{willems2005note}'s Fundamental Lemma states:
\begin{equation}
    \mathcal B_L=\im\mathcal H_L(\mbf w^{\mrm d}).
    \label{eq:preliminaries:fundamental}
\end{equation}

For prediction, choose a history length \(T_{\mrm{ini}}\geq\ell\) and a prediction horizon \(N\), and set \(L=T_{\mrm{ini}}+N\). In every column of a Hankel matrix, the first \(T_{\mrm{ini}}\) samples form the past (superscript \(\mrm p\)) and the remaining \(N\) samples form the future (superscript \(\mrm f\)). The input and output Hankel matrices are therefore \(H_u=\mathcal H_L(\mbf u^{\mrm d})=\col(U^{\mrm p},U^{\mrm f})\) and \(H_y=\mathcal H_L(\mbf y^{\mrm d})=\col(Y^{\mrm p},Y^{\mrm f})\). With measured outputs, the output matrix becomes \(\wt H_y=\mathcal H_L(\wt{\mbf y}^{\mrm d})=\col(\wt Y^{\mrm p},\wt Y^{\mrm f})\). All blocks have \(n_g\) columns, with \(n_g=T_{\mrm d}-L+1\) for a single record. The blocks \(U^{\mrm p}\) and \(U^{\mrm f}\) have \(n_uT_{\mrm{ini}}\) and \(n_uN\) rows, and the output blocks have \(n_yT_{\mrm{ini}}\) and \(n_yN\) rows.

Define the histories \(\mbf u_k^{\mrm{ini}}=\mbf u_{k-T_{\mrm{ini}}:k-1}\) and \(\mbf y_k^{\mrm{ini}}=\mbf y_{k-T_{\mrm{ini}}:k-1}\), and let \(\wt{\mbf y}_k^{\mrm{ini}}\) denote the measured output history. Under the noiseless spanning conditions, \((\mbf u_k,\mbf y_k)\) is a valid continuation of this history if and only if some \(g_k\in\R[n_g]\) satisfies
\begin{equation}
    \col(U^{\mrm p},Y^{\mrm p},U^{\mrm f},Y^{\mrm f})g_k
    =\col(\mbf u_k^{\mrm{ini}},\mbf y_k^{\mrm{ini}},\mbf u_k,\mbf y_k).
    \label{eq:preliminaries:data-equation}
\end{equation}
Since \(T_{\mrm{ini}}\geq\ell\), the history determines the initial state, and the prediction is equivalent to model-based prediction with the same cost and constraints~\citep{coulson2019data}.

\subsection{Regularised DeePC}
\label{sec:preliminaries:deepc}

Measurement noise perturbs both the recorded matrix and the online history, so~\eqref{eq:preliminaries:data-equation} cannot be satisfied exactly. Regularised DeePC therefore allows a residual \(\sigma_k^y\) in the past-output equation and penalises the coefficient vector \(g_k\)~\citep{coulson2019regularized}. Penalising \(\norm{g_k}_2\) pulls the prediction towards the zero trajectory, which is unsuitable for a nonzero reference. Following \citet{elokda2021data}, we centre the penalty at a coefficient vector that reproduces the reference. Define
\begin{equation}
    \begin{aligned}
        \wt H^{\mrm d} & =\col(U^{\mrm p},\wt Y^{\mrm p},U^{\mrm f},\wt Y^{\mrm f}),\qquad
        g_k^{\mrm{ref}}=(\wt H^{\mrm d})^\dagger r_k,                                                             \\
        r_k            & =\col(\mbf u^{\mrm{ref}}_{k-T_{\mrm{ini}}:k-1},\mbf y^{\mrm{ref}}_{k-T_{\mrm{ini}}:k-1},
        \mbf u^{\mrm{ref}}_{k:k+N-1},\mbf y^{\mrm{ref}}_{k:k+N-1}).
    \end{aligned}
    \label{eq:preliminaries:coefficient-reference}
\end{equation}
The minimum-norm least-squares solution \(g_k^{\mrm{ref}}\) is recomputed when the reference or the data change. It reproduces \(r_k\) exactly if \(r_k\in\im\wt H^{\mrm d}\), which does not imply physical feasibility.

The regularised problem is
\begin{equation}
    \begin{aligned}
        \min_{g_k,\,\sigma_k^y}\quad & J_k(U^{\mrm f}g_k,\wt Y^{\mrm f}g_k)
        +\lambda_y\norm{\sigma_k^y}_2^2
        +\lambda_g\norm{g_k-g_k^{\mrm{ref}}}_2^2                                 \\
        \text{subject to}\quad       & U^{\mrm p}g_k=\mbf u_k^{\mrm{ini}},\qquad
        \wt Y^{\mrm p}g_k=\wt{\mbf y}_k^{\mrm{ini}}+\sigma_k^y,
    \end{aligned}
    \label{eq:preliminaries:regularised-deepc}
\end{equation}
where \(\lambda_y,\lambda_g>0\). Known inputs \(\mbf u_k^{\mrm{ini}}\) are matched exactly, and the penalised past-output residual is \(\sigma_k^y=\wt Y^{\mrm p}g_k-\wt{\mbf y}_k^{\mrm{ini}}\). The future predictions are \(\mbf u_k=U^{\mrm f}g_k\) and \(\mbf y_k=\wt Y^{\mrm f}g_k\).

\subsection{Geometric Robust DDPC}
\label{sec:preliminaries:geometric}

Geometric robust DDPC~\citep{bharadwaj2026robust} accounts for uncertainty in the subspace that represents the trajectories. Let \(w_j=\col(u_j,\wt y_j)\) and \(n_{\mrm{amb}}=(n_u+n_y)L\). The leading \(d<n_{\mrm{amb}}\) left singular vectors of \(\mathcal H_L(\wt{\mbf w}^{\mrm d})\) form an orthonormal basis \(\wh V\) of the fitted subspace \(\wh{\mathcal S}=\im\wh V\). The dimension \(d\) is motivated by the LTI behaviour dimension \(n_uL+n_x\). Nonlinear effects and measurement noise raise the rank of the data matrix, \(\wh{\mathcal S}\) becomes a fixed-rank approximation. The data may be centred and scaled before fitting (\Cref{app:geometric:coordinates}).

The Grassmannian \(\Gr(d,n_{\mrm{amb}})\) is the set of \(d\)-dimensional subspaces of \(\R[n_{\mrm{amb}}]\). For an orthonormal basis \(V\) of \(\mathcal S\), the projector is \(P_{\mathcal S}=VV^\top\). The principal angles \(\theta_j\in[0,\pi/2]\) between two subspaces satisfy \(\cos\theta_j=\sigma_j(V_1^\top V_2)\), where \(\sigma_j\) is the \(j\)th singular value. The chordal distance and the open ball of radius \(\rho\) around \(\wh{\mathcal S}\) are
\begin{equation}
    \begin{aligned}
        d_{\mrm{ch}}(\mathcal S_1,\mathcal S_2)
         & =\frac{\norm{P_{\mathcal S_1}-P_{\mathcal S_2}}_{\mrm F}}{\sqrt2}
        =\left(\sum_{j=1}^{d}\sin^2\theta_j\right)^{1/2},                                                       \\
        \mathbb B_\rho(\wh{\mathcal S})
         & =\{\mathcal S\in\Gr(d,n_{\mrm{amb}}):d_{\mrm{ch}}(\mathcal S,\wh{\mathcal S})<\rho\},\quad 0<\rho<1.
    \end{aligned}
    \label{eq:preliminaries:ball}
\end{equation}

\begin{figure}[ht!]
    \centering
\begin{tikzpicture}[>=Latex,font=\small,x=0.72cm,y=0.72cm]
  \def\Rad{5.0}\def\rhoang{24}\def\advang{-15}
  \coordinate (O) at (0,0);
  \path[fill=blue!9] (\rhoang:\Rad)--(O)--(-\rhoang:\Rad)--cycle;
  \path[fill=blue!9] ({180-\rhoang}:\Rad)--(O)--({180+\rhoang}:\Rad)--cycle;
  \draw[blue!50!black,dashed] (\rhoang:\Rad)--({180+\rhoang}:\Rad);
  \draw[blue!50!black,dashed] (-\rhoang:\Rad)--({180-\rhoang}:\Rad);
  \draw[very thick,blue!65!black] (-\Rad,0)--(\Rad,0) node[right]{\(\widehat{\mathcal S}\)};
  \draw[very thick,orange!80!black] ({180+\advang}:\Rad)--(\advang:\Rad) node[right]{\(\mathcal S^{\mathrm a}\)};
  \coordinate (xi) at (3.5,3.1);
  \coordinate (b)  at (4.1,1.3);
  \coordinate (Pn) at ($(O)!(xi)!(\Rad,0)$);
  \coordinate (Pa) at ($(O)!(xi)!(\advang:\Rad)$);
  \draw[dotted] (xi)--(Pn); \draw[dotted] (xi)--(Pa);
  \draw[->,thick] (O)--(xi) node[below right=1pt]{\(\xi\)};
  \fill (Pn) circle(2pt) node[below right=1pt]{\(P_{\widehat{\mathcal S}}\,\xi\)};
  \fill (Pa) circle(2pt) node[below left=3pt and 0pt]{\(P_{\mathcal S^{\mathrm a}}\xi\)};
  \fill[red!70!black] (b) circle(2.2pt) node[right=2pt]{\(b_k\)};
  \draw[blue!65!black,thick] (Pn)--(b);
  \draw[orange!80!black,thick] (Pa)--(b);
  \draw[->] (1.2,0) arc[start angle=0,end angle=\rhoang,radius=1.2];
  \node[font=\footnotesize] at (1.55,0.32) {\(\theta_\rho\)};
  \node[blue!50!black,font=\footnotesize,anchor=east] at (-1.3,1.45) {\(\mathbb B_\rho(\widehat{\mathcal S})\)};
  \node[font=\footnotesize,anchor=east] at (-2.8,-1.1) {\(\rho=\sin\theta_\rho\)};
\end{tikzpicture}
    \caption{Geometric uncertainty in \(\Gr(1,2)\). The fitted subspace \(\wh{\mathcal S}\) is the horizontal line. The shaded double cone is the ball \(\mathbb B_\rho(\wh{\mathcal S})\) of lines with \(|\theta|<\theta_\rho=\sin^{-1}\rho\), (dashed boundary lines are excluded). Projecting \(\xi\) onto \(\wh{\mathcal S}\) and onto an adverse subspace \(\mathcal S^{\mrm a}\) gives different trajectories, the adverse with the larger residual to \(b_k\).}
    \label{fig:preliminaries:ball}
\end{figure}
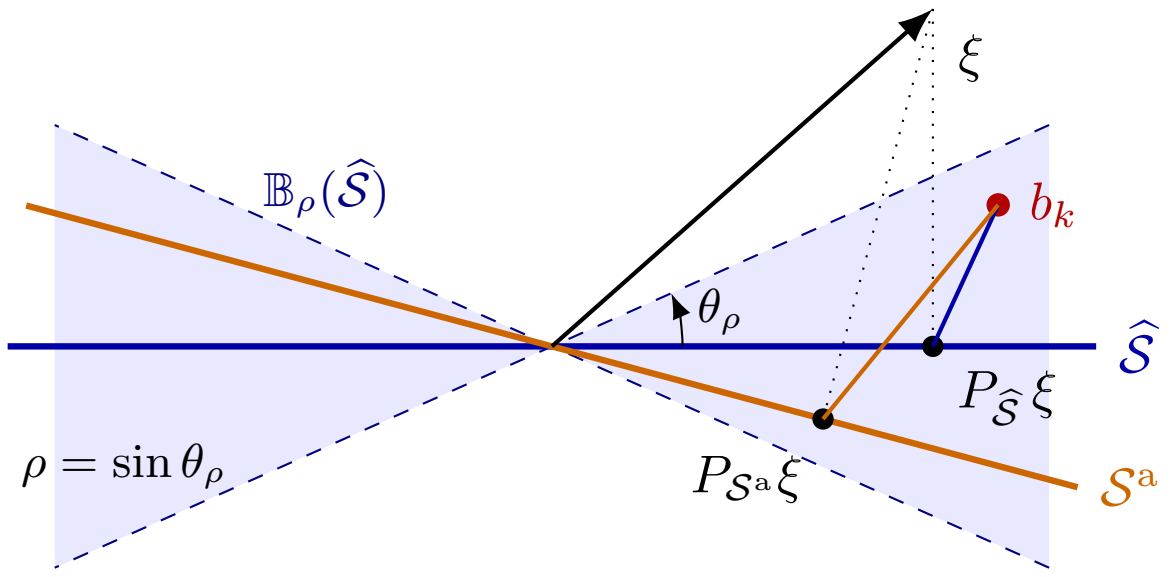

At update \(k\), the target \(b_k\in\R[n_{\mrm{amb}}]\) collects the measured past pairs \(\col(u_j,\wt y_j)\) for \(j=k-T_{\mrm{ini}},\ldots,k-1\) and the future reference pairs \(\col(u_j^{\mrm{ref}},y_j^{\mrm{ref}})\) for \(j=k,\ldots,k+N-1\). For a decision variable \(\xi\in\R[n_{\mrm{amb}}]\), the projection \(P_{\mathcal S}\xi\) onto a candidate subspace \(\mathcal S\) is the corresponding predicted trajectory, with past and future inputs and outputs. The predictive problem is
\begin{equation}
    \min_{\xi\in\R[n_{\mrm{amb}}]}\\
    \sup_{\mathcal S\in\mathbb B_\rho(\wh{\mathcal S})}F_k(\xi,\mathcal S),\qquad
    F_k(\xi,\mathcal S)=\norm{P_{\mathcal S}\xi-b_k}_W^2,
    \label{eq:preliminaries:geometric}
\end{equation}
with \(W=\diag\bigl((1+\gamma)I_{(n_u+n_y)T_{\mrm{ini}}},I_{(n_u+n_y)N}\bigr)\) and \(\gamma>0\). The past block of \(W\) penalises deviations from the measured history, with \(\gamma\) controlling how strongly the prediction is tied to it, and the future block penalises tracking errors. The radius \(\rho\) sets the subspace uncertainty. \Cref{fig:preliminaries:ball} illustrates the projections.

The objective in~\eqref{eq:preliminaries:geometric} is convex in \(\xi\) as a supremum of convex functions, while the search for the adverse subspace is nonconvex. Because \(F_k\) is continuous in \(\mathcal S\), the supremum over the open ball equals the maximum over its closure. For the returned \(\xi\), the adverse subspace \(\mathcal S^{\mrm a}\) maximises \(F_k(\xi,\cdot)\) over the ball. The controller decodes the trajectory \(P_{\mathcal S^{\mrm a}}\xi\) and applies its first future input. Both inputs and outputs vary with the subspace. The ball also contains subspaces without a causal LTI realisation, so its use here is evaluated experimentally.

\section{Planar 2R Manipulator Control}
\label{sec:manipulator}

The mechanical model feedback-linearises the arm, and geometric DDPC predicts the resulting virtual-input--position response.

\subsection{Manipulator Model and Tracking Task}
\label{sec:manipulator:model}

\begin{minipage}[c]{0.64\linewidth}
    The base angle \(q_1\) is measured from the positive horizontal axis; \(q_2\) is relative to link~1, so link~2 has absolute angle \(q_1+q_2\). Link \(j\) has length \(l_j\), mass \(m_j\), centre-of-mass distance \(r_j\) from its proximal joint, and inertia \(\mathcal I_j\) about its centre of mass. \Cref{fig:manipulator:geometry} shows this convention and the centre-of-mass positions \(p_{c1},p_{c2}\). For uniform rods, \(r_j=l_j/2\) and \(\mathcal I_j=m_jl_j^2/12\).
\end{minipage}
\hfill
\begin{minipage}[c]{0.32\linewidth}
    \centering
    \resizebox{\linewidth}{!}{
\begin{tikzpicture}[>=Latex,scale=0.76,font=\small]
 \coordinate(O) at (0,0); \coordinate(J) at (3.2,1.8); \coordinate(T) at (4.5,4.2);
 \draw[->,thin] (O)--(2,0) node[right]{\(x\)};
 \draw[->,thin] (O)--(0,2) node[above]{\(y\)};
 \draw[very thick,blue!65!black] (O)--(J);
 \draw[very thick,orange!75!black] (J)--(T);
 \node[blue!65!black,anchor=north west] at (2.45,1.38) {\(l_1,m_1,\mathcal I_1\)};
 \node[orange!75!black,anchor=west] at (4.18,3.62) {\(l_2,m_2,\mathcal I_2\)};
 \draw[dashed] (J)--++(1.35,.76);
 \draw[->] (.75,0) arc[start angle=0,end angle=29.36,radius=.75];
 \node at (.98,.2){\(q_1\)};
 \draw[->] (4.05,2.278) arc[start angle=29.36,end angle=61.56,radius=.975];
 \node at (3.66,2.2){\(q_2\)};
 \draw[->,thick] (-0.72,2.7)--(-0.72,1.35) node[midway,left]{$g_0$};
 \foreach \p in {O,J,T}{\fill (\p) circle(2.1pt);}
 \coordinate(C1) at (1.6,.9); \coordinate(C2) at (3.85,3.0);
 \draw[fill=white] (C1) circle(2.4pt); \draw[fill=white] (C2) circle(2.4pt);
 \node[below right=1pt] at (C1){\(p_{c1}\)}; \node[right=2pt] at (C2){\(p_{c2}\)};
 \draw[<->,thin] (-.22,.39)--node[above left=-1pt]{\(r_1\)} (1.38,1.29);
 \draw[<->,thin] (2.85,1.99)--node[left=-1pt]{\(r_2\)} (3.5,3.19);
 \draw[thin] (-.22,-.13)--(.22,-.13);
 \foreach \a in {-.16,0,.16}{\draw[thin] (\a,-.13)--++(-.10,-.14);}
\end{tikzpicture}
}
    
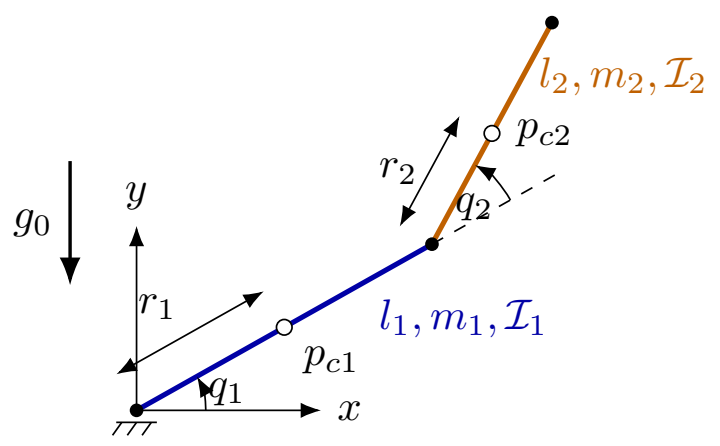
\captionof{figure}{Planar 2R Manipulator.}
    \label{fig:manipulator:geometry}
\end{minipage}

Let \(q=\col(q_1,q_2)\), \(v=\dot q\), and \(x=\col(q,v)\). The joint dynamics are
\begin{equation}
    M(q)\ddot q+\eta(q,v)+g(q)+D_{\mrm v}v=\tau^{\mrm{act}},
    \label{eq:manipulator:dynamics}
\end{equation}
where \(M(q)\succ0\) is the inertia matrix, \(\eta\) contains Coriolis and centrifugal terms, \(g\) is the gravity torque, \(D_{\mrm v}\succeq0\) is diagonal viscous damping, and \(\tau^{\mrm{act}}\) is the applied torque. The expressions for \(M\) and \(\eta\) are given in \Cref{app:estimation}.
For gravitational acceleration of magnitude \(g_0\) in the negative vertical direction,
\begin{equation}
    g(q)=g_0\begin{bmatrix}
        (m_1r_1+m_2l_1)\cos q_1+m_2r_2\cos(q_1+q_2) \\
        m_2r_2\cos(q_1+q_2)
    \end{bmatrix}.
    \label{eq:manipulator:gravity}
\end{equation}

Positions are measured at \(t_k=kT_{\mrm s}\) as \(\wt y_k=q(t_k)+\varepsilon_k^y\), with independent \(\varepsilon_k^y\sim\mathcal N(0,\Sigma_y)\). Velocity is estimated. The task is to track \(q^{\mrm{ref}}(t)\), assessed by position error and applied torque. Predictive problems impose no joint or torque constraints.

\subsection{Feedback Linearisation and State Estimation}
\label{sec:manipulator:compensation}

Write \(h(q,v)=\eta(q,v)+g(q)+D_{\mrm v}v\). The supplied model uses \(\wh M(q)\) and \(\wh h(q,v)\), and its parameters may differ from the plant. For a held virtual acceleration \(\mu_k\in\R[2]\), the inverse dynamics map gives
\begin{equation}
    \tau^{\mrm{id}}(t)=\wh M(\wh q(t))\mu_k+\wh h(\wh q(t),\wh v(t)).
    \label{eq:manipulator:inverse-dynamics}
\end{equation}
With exact parameters, state feedback, continuous torque evaluation and inactive actuator limits, substitution gives \(\ddot q=\mu_k\). The feedback-linearised system then consists of two double integrators, with exact sampled model \(x_{k+1}=Ax_k+B\mu_k\), \(y_k=Cx_k\), where
\begin{equation}
    A=\begin{bmatrix}I_2&T_{\mrm s}I_2\\0&I_2\end{bmatrix},\qquad
    B=\begin{bmatrix}\tfrac12T_{\mrm s}^2I_2\\T_{\mrm s}I_2\end{bmatrix},\qquad
    C=\begin{bmatrix}I_2&0\end{bmatrix}.
    \label{eq:manipulator:virtual-lti}
\end{equation}
The ideal model is controllable, has observability lag two, and has \(\dim\mathcal B_L = 2L+4\) for \(L\geq2\).

A Kalman filter estimates position and velocity from measured positions and commanded virtual inputs~\citep{welch2000introduction}, using~\eqref{eq:manipulator:virtual-lti}. \Cref{fig:manipulator:architecture} shows the loop; \Cref{app:estimation} gives the filter.

The difference between the actual and continuously evaluated model torque is \(\Delta\tau^{\mrm{act}}=\tau^{\mrm{act}}-\tau^{\mrm{id}}\). This includes torque holding and actuator limitations. Consequently, \(\ddot q=\mu_k+a^{\mrm{res}}\), with
\begin{equation}
    a^{\mrm{res}}=M(q)^{-1}\!\left[
        (\wh M(\wh q)-M(q))\mu_k+
        \wh h(\wh q,\wh v)-h(q,v)+\Delta\tau^{\mrm{act}}\right].
    \label{eq:manipulator:residual}
\end{equation}
With exact state feedback and unrestricted continuous torque application, the dynamics become \(\ddot q=G(q)\mu_k+f_{\mrm r}(q,v)\), where \(G=M^{-1}\wh M\) and \(f_{\mrm r}=M^{-1}(\wh h-h)\). The recorded response contains these model errors together with estimation and actuator effects, giving the learned predictor information beyond the ideal double integrator.

\subsection{Offline Data and Reference Initialisation}
\label{sec:manipulator:data}

A PD tracking controller with acceleration feedforward and random excitation generates offline data. Its sinusoidal reference \(q^{\mrm{col}}(t)\) varies in centre, amplitude, phase and period between episodes (\Cref{app:data}).

For the feedback-linearised arm, \(u_k=\mu_k\) is held for \(T_{\mrm s}\), while inverse dynamics is evaluated from the propagated estimate every \(T_\tau\). For direct data, \(u_k=\tau_k^{\mrm{cmd}}\) is computed once at \(t_k\) and held for \(T_{\mrm s}\). The actuator enforces torque magnitude and slew limits at each \(T_\tau\) update in both cases. The mechanical model and state estimates are used to generate both offline datasets. Windows are concatenated without crossing episode boundaries.

Tracking references are joint sinusoids with a smooth entry ramp. Sampled \(q_k^{\mrm{ref}},v_k^{\mrm{ref}},\mu_k^{\mrm{ref}}\) satisfy the ideal dynamics through a piecewise-constant-acceleration interpolant. Torque and torque-rate requirements are checked throughout the motion (\Cref{app:data:references}).

\subsection{Geometric DDPC for the Feedback-Linearised Arm}
\label{sec:manipulator:geometric}

A rank-\(d=2L+4\) fit to records of \((\mu_k,\wt y_k)\) defines \(\wh{\mathcal S}\). This dimension follows the ideal four-state model; the affine fitting coordinates are specified in \Cref{app:geometric:coordinates}.

At update \(k\), the target \(b_k\) contains the preceding \(T_{\mrm{ini}}\) measured pairs and the next \(N\) reference pairs \((\mu_j^{\mrm{ref}},q_j^{\mrm{ref}})\), expressed in the predictor's coordinates. The controller solves~\eqref{eq:preliminaries:geometric} over the open ball in \(\Gr(2L+4,4L)\), with fixed \(\rho\) and \(\gamma\). A scenario-exchange algorithm returns an approximate adverse trajectory, whose first future virtual input is applied through~\eqref{eq:manipulator:inverse-dynamics}. \Cref{app:geometric:solver} gives the solve and stopping rules.

\begin{figure}[ht!]
    \centering
    \begin{minipage}[t]{0.48\textwidth}
        \centering
        \resizebox{\linewidth}{!}{
\begin{tikzpicture}[>=Latex,font=\small,x=1cm,y=1cm,
  box/.style={draw,align=center,inner sep=3pt,minimum height=8mm},
  arr/.style={->,line width=.55pt}]
  \node[box,text width=1.55cm] (ddpc) at (0,0) {Geometric\\DDPC};
  \node[box,text width=1.65cm] (id) at (3.15,0) {Inverse\\dynamics};
  \node[box,text width=1.55cm] (plant) at (6.45,0) {Actuator\\and arm};
  \node[box,text width=1.65cm] (observer) at (3.15,-1.4) {Observer};
  \node[box,text width=1.55cm] (history) at (0,-1.4) {Past pairs};
  \draw[arr] (ddpc.east)--node[above]{\(\mu_k\)} (id.west);
  \draw[arr] (id.east)--node[above]{\(\tau^{\mrm{cmd}}\)} (plant.west);
  \draw[arr] (observer.north)--node[right,inner sep=2pt]{\(\wh x(t)\)} (id.south);
  \coordinate (command) at (1.45,0);
  \draw[arr] (command)--(1.45,-1.4)--(observer.west);
  \draw[arr] (1.45,-1.4)--(history.east);
  \fill (command) circle(1pt);
  \fill (1.45,-1.4) circle(1pt);
  \draw[arr] (plant.east)--(8,0)--(8,-1.4)--node[above]{\(\wt y_k\)}(observer.east);
  \draw[arr] (8,-1.4)--(8,-2.05)--(0,-2.05)--(history.south);
  \fill (8,-1.4) circle(1pt);
  \draw[arr] (history.north)--(ddpc.south);
  \draw[arr] (-1.85,0) node[left]{reference}--(ddpc.west);
\end{tikzpicture}}
        \caption{Geometric DDPC with feedback linearisation. The observer uses the current position measurement. The predictor uses completed \((\mu_j,\wt y_j)\) pairs through \(j=k-1\).}
        \label{fig:manipulator:architecture}
    \end{minipage}
    \hfill
    \begin{minipage}[t]{0.48\textwidth}
        \centering
        \resizebox{\linewidth}{!}{\begin{tikzpicture}[>=Latex,font=\small,x=1cm,y=1cm,
  box/.style={draw,align=center,inner sep=3pt,minimum height=8mm},
  arr/.style={->,line width=.55pt}]
  \node[box,text width=1.55cm] (deepc) at (0,0) {Direct\\DeePC};
  \node[box,text width=1.55cm] (plant) at (6.45,0) {Actuator\\and arm};
  \node[box,text width=2.5cm] (history) at (3.15,-1.05) {Past pairs and\\online data bank};
  \draw[arr] (deepc.east)--node[above]{\(\tau_k^{\mrm{cmd}}\)} (plant.west);
  \draw[arr] (3.15,0)--(history.north);
  \fill (3.15,0) circle(1pt);
  \draw[arr] (plant.east)--(8,0)--(8,-1.05)--node[above]{\(\wt y_k\)} (history.east);
  \draw[arr] (history.west)--(0,-1.05)--(deepc.south);
  \draw[arr] (-1.85,0) node[left]{reference}--(deepc.west);
\end{tikzpicture}}
        \caption{Direct regularised DeePC. The desired torque and position sequences define the reference, while completed torque--position pairs update the history and online data bank.}
        \label{fig:manipulator:direct}
    \end{minipage}
\end{figure}

\subsection{Comparison Controllers}
\label{sec:manipulator:comparators}

\paragraph{Double-integrator MPC.}
This controller predicts with~\eqref{eq:manipulator:virtual-lti}, using the same horizon, observer, feedback-linearising torque map and actuator as geometric DDPC. It minimises~\eqref{eq:preliminaries:tracking-cost} with \(u=\mu\) and \(y=q\), using the same future tracking weights as geometric DDPC in physical coordinates. Its first virtual input is applied through the shared torque map.

\paragraph{Direct regularised DeePC.}
Direct DeePC uses the recorded torque and position pairs to predict future motion~\citep{elokda2021data,wegner2021data}. In \Cref{fig:manipulator:direct}, the desired motion defines the torque reference \(\tau_k^{\mrm{ref}}=\wh M(q_k^{\mrm{ref}})\mu_k^{\mrm{ref}}+\wh h(q_k^{\mrm{ref}},v_k^{\mrm{ref}})\). Set \(u^{\mrm{ref}}=\tau^{\mrm{ref}}\) and \(y^{\mrm{ref}}=q^{\mrm{ref}}\) in both the tracking cost and \(g^{\mrm{ref}}\)~\eqref{eq:preliminaries:coefficient-reference}. At each update, the measured history initialises~\eqref{eq:preliminaries:regularised-deepc}, whose first predicted torque is applied directly. The model supplies references and collection torques; online feedback uses measured input--output history. \Cref{app:comparators:direct} gives the quadratic solve.

\section{Simulation Studies}
\label{sec:simulation}

We examine prediction accuracy (A), transfer of predictive tuning (B), and cyclic tracking (C). Controllers receive noisy positions; true positions are used only to evaluate errors.

\subsection{Simulation Setup and Evaluation}
\label{sec:simulation:setup}

\Cref{tab:simulation:arms} defines the mechanical conditions. Experiments~A and~C retain a fixed model while changing the plant parameters. Experiment~B uses exact models of two known arms. In all cases, gravity acts downward and position noise has covariance \(\Sigma_y=\sigma_y^2I_2\) during collection and control.

\begin{table}[ht!]
    \centering\small
    \setlength{\tabcolsep}{5pt}
    \begin{tabular}{@{}lllll@{}}
        \toprule
        Condition        & Studies & Lengths [m]     & True masses [kg] & Model masses [kg] \\
        \midrule
        Exact parameters & A, C    & \((1.0,0.8)\)   & \((0.9,0.72)\)   & \((0.9,0.72)\)    \\
        Parameter errors & A, C    & \((1.0,0.8)\)   & \((1.0,0.8)\)    & \((0.9,0.72)\)    \\
        Arm A            & B       & \((1.0,0.8)\)   & \((1.0,0.8)\)    & \((1.0,0.8)\)     \\
        Arm B            & B       & \((0.75,0.55)\) & \((1.3,0.6)\)    & \((1.3,0.6)\)     \\
        \bottomrule
    \end{tabular}
    \caption{Mechanical conditions. Model lengths equal true lengths. True damping is \((0.04,0.02)\,\mathrm{N\,m\,s\,rad^{-1}}\) in the parameter-error condition and zero elsewhere; model damping is zero throughout.}
    \label{tab:simulation:arms}
\end{table}

Controller settings are fixed before evaluation. Direct-DeePC weights are selected on separate Arm~A trials at baseline noise. The data-driven methods share the collection procedure and budget, with their respective input conventions. Trials use paired measurement-noise sequences across controllers, and predictive computation is instantaneous. Each run covers a smooth entry ramp and two reference cycles. \Cref{app:simulation} gives the settings and evaluation protocol.

Tables report median [minimum, maximum] periodic position RMSE \(E_{\mrm{cyc}}\), median whole-run applied-torque RMS \(\tau_{\mrm{rms}}\), and completed trials \(n_{\mrm c}\). Both measures average over the two joints. Incomplete trials enter completion counts but are excluded from full-duration error and effort statistics.

\subsection{Prediction Accuracy After Feedback Linearisation}
\label{sec:simulation:prediction}

Experiment~A compares the double integrator with the predictor fitted to virtual-input--position data. On independent test trajectories, both predict future positions from the same prescribed inputs and past measured positions~\citep{markovsky2008data}. Future positions enter only the error calculation. \Cref{tab:simulation:prediction} covers exact and inaccurate mechanics at three noise levels.

\begin{table}[ht!]
    \centering\small
    \setlength{\tabcolsep}{4pt}
    \renewcommand{\arraystretch}{1.12}
    \begin{tabular}{@{}lrrr@{}}
        \toprule
        Mechanics        & \(\sigma_y\) [rad] & Double integrator       & Learned behaviour       \\
        \midrule
        Exact parameters & 0.001              & 0.0098 [0.0068, 0.0110] & 0.0058 [0.0051, 0.0070] \\
                         & 0.005              & 0.0157 [0.0118, 0.0191] & 0.0129 [0.0094, 0.0147] \\
                         & 0.010              & 0.0251 [0.0213, 0.0279] & 0.0200 [0.0191, 0.0242] \\
        \midrule
        Parameter errors & 0.001              & 0.2096 [0.1831, 0.2496] & 0.0153 [0.0131, 0.0194] \\
                         & 0.005              & 0.1801 [0.1510, 0.2207] & 0.0167 [0.0104, 0.0183] \\
                         & 0.010              & 0.1598 [0.1292, 0.1916] & 0.0234 [0.0187, 0.0270] \\
        \bottomrule
    \end{tabular}
    \caption{Position prediction RMSE [rad]. Entries are medians [minimum, maximum] across independently fitted models, after taking the median error over each model's test windows.}
    \label{tab:simulation:prediction}
\end{table}

Learning gives a modest improvement with exact parameters and a much larger improvement with parameter errors, indicating that data capture systematic departures from the ideal dynamics. The learned predictor remains more accurate at each noise level, although its error grows with noise. Noise also changes the feedback-controlled collection trajectories, so comparisons across noise levels include this effect.

\subsection{Transfer Across Noise and Arm Changes}
\label{sec:simulation:transfer}

Experiment~B uses the base cycle with unchanged predictive tuning. Noise is increased on Arm~A, then a known Arm~B with shorter links and different masses is tested. Each condition uses newly collected data and the stated observer noise covariance; the mechanical model is updated for Arm~B. The collection procedure and predictive settings stay fixed.

\begin{table}[ht!]
    \centering\small
    \setlength{\tabcolsep}{3pt}
    \renewcommand{\arraystretch}{1.12}
    \begin{tabular}{@{}llrrr@{}}
        \toprule
        Condition                 & Controller     & \(E_{\mrm{cyc}}\)       & \(\tau_{\mrm{rms}}\) & \(n_{\mrm c}/6\) \\
        \midrule
        Arm A, \(\sigma_y=0.001\) & Geometric DDPC & 0.0042 [0.0038, 0.0054] & 2.984                & 6/6              \\
                                  & Direct DeePC   & 0.0245 [0.0236, 0.0266] & 2.961                & 6/6              \\
        \midrule
        Arm A, \(\sigma_y=0.005\) & Geometric DDPC & 0.0138 [0.0116, 0.0164] & 2.977                & 6/6              \\
                                  & Direct DeePC   & 0.0370 [0.0285, 0.0545] & 2.955                & 6/6              \\
        \midrule
        Arm A, \(\sigma_y=0.010\) & Geometric DDPC & 0.0280 [0.0210, 0.0345] & 3.560                & 6/6              \\
                                  & Direct DeePC   & 0.0662 [0.0369, 0.2517] & 2.949                & 5/6              \\
        \midrule
        Arm B, \(\sigma_y=0.001\) & Geometric DDPC & 0.0048 [0.0039, 0.0055] & 1.957                & 6/6              \\
                                  & Direct DeePC   & 0.0257 [0.0227, 0.0323] & 1.948                & 6/6              \\
        \bottomrule
    \end{tabular}
    \caption{Transfer with fixed predictive settings. Errors are in radians and torque RMS in \(\mathrm{N\,m}\); \(n_{\mrm c}/6\) gives completed/attempted trials.}
    \label{tab:simulation:transfer}
\end{table}

\begin{figure}[ht!]
    \centering
    \includegraphics[width=0.49\linewidth]{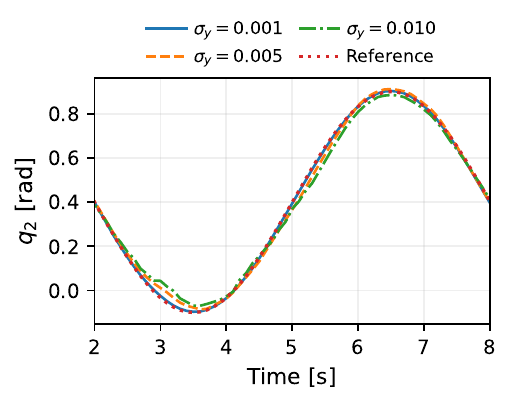}\hfill
    \includegraphics[width=0.49\linewidth]{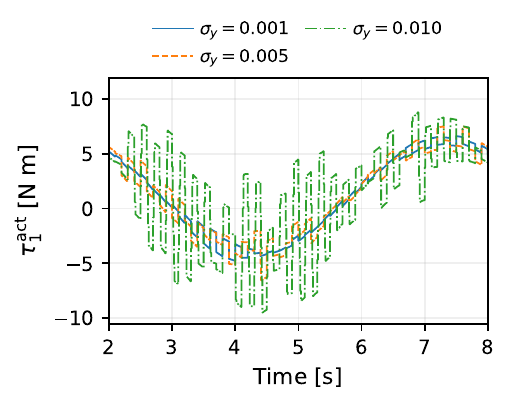}
    \caption{Noise transfer with geometric DDPC on Arm~A: second-joint position and applied first-joint torque in the first trial, after the ramp. Legend values give \(\sigma_y\) in radians.}
    \label{fig:simulation:noise-transfer}
\end{figure}

Tracking error increases with noise for both methods (\Cref{tab:simulation:transfer}). Geometric DDPC retains lower median error and completes every trial. At high noise, its torque variation and actuator activity increase (\Cref{fig:simulation:noise-transfer}); direct DeePC has a wider error spread and one incomplete trial. On Arm~B, both methods retain accuracy close to their Arm~A baselines with reduced torque effort. Geometric DDPC remains more accurate at similar effort.

\subsection{Trajectory Tracking}
\label{sec:simulation:tracking}

Experiment~C compares geometric DDPC, double-integrator MPC and direct regularised DeePC. The base reference is tested with exact mechanical parameters; all three references in \Cref{tab:simulation:references} are tested with parameter errors. The mechanical model stays fixed, and offline data are reused across references within each plant condition. Sampled reference torques satisfy the actuator limits.

\begin{table}[ht!]
    \centering\small
    \begin{tabular}{@{}llll@{}}
        \toprule
        Reference & Centre [rad]           & Amplitudes [rad] & Period [s] \\
        \midrule
        Base      & \((-\pi/2,0.4)\)       & \((0.35,0.50)\)  & 6          \\
        Shifted   & \((-\pi/2+0.05,0.48)\) & \((0.42,0.58)\)  & 6          \\
        Faster    & \((-\pi/2,0.4)\)       & \((0.30,0.42)\)  & 4.5        \\
        \bottomrule
    \end{tabular}
    \caption{Joint references, with phases \((0,\pi/3)\) and a two-second entry ramp.}
    \label{tab:simulation:references}
\end{table}

\begin{figure}[ht!]
    \centering
    \includegraphics[width=0.49\linewidth]{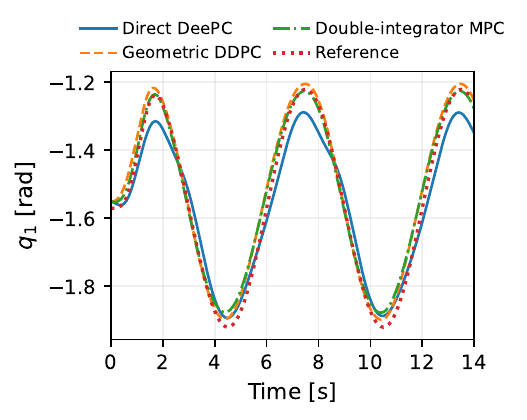}\hfill
    \includegraphics[width=0.49\linewidth]{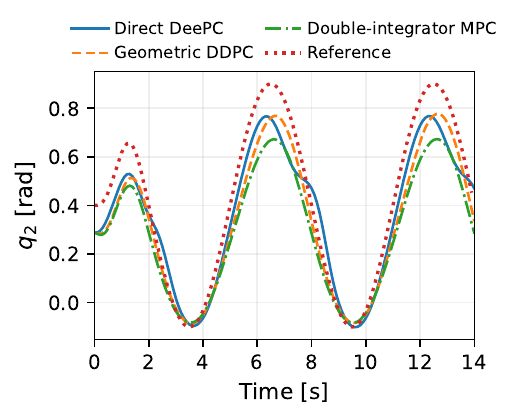}
    \caption{Base-cycle tracking with mass and damping errors. Both joint positions are shown for the first trial of each controller, together with the reference.}
    \label{fig:simulation:tracking}
\end{figure}

\Cref{tab:simulation:tracking} summarises tracking accuracy and effort.
\Cref{fig:simulation:tracking} shows the base-cycle trajectories with parameter errors.

\begin{table}[ht!]
    \centering\small
    \setlength{\tabcolsep}{3pt}
    \renewcommand{\arraystretch}{1.12}
    \begin{tabular}{@{}llrrr@{}}
        \toprule
        Condition, cycle          & Controller            & \(E_{\mrm{cyc}}\)       & \(\tau_{\mrm{rms}}\) & \(n_{\mrm c}/6\) \\
        \midrule
        Exact parameters, base    & Double-integrator MPC & 0.0050 [0.0046, 0.0053] & 2.703                & 6/6              \\
                                  & Geometric DDPC        & 0.0050 [0.0047, 0.0064] & 2.690                & 6/6              \\
                                  & Direct DeePC          & 0.0253 [0.0213, 0.0292] & 2.665                & 6/6              \\
        \midrule
        Parameter errors, base    & Double-integrator MPC & 0.1091 [0.1090, 0.1093] & 2.810                & 6/6              \\
                                  & Geometric DDPC        & 0.0650 [0.0482, 0.0749] & 3.030                & 6/6              \\
                                  & Direct DeePC          & 0.0826 [0.0636, 0.0848] & 2.530                & 6/6              \\
        \midrule
        Parameter errors, shifted & Double-integrator MPC & 0.1152 [0.1150, 0.1154] & 3.456                & 6/6              \\
                                  & Geometric DDPC        & 0.0592 [0.0412, 0.0744] & 3.681                & 6/6              \\
                                  & Direct DeePC          & 0.1365 [0.1089, 0.1411] & 3.043                & 6/6              \\
        \midrule
        Parameter errors, faster  & Double-integrator MPC & 0.1068 [0.1068, 0.1069] & 2.209                & 6/6              \\
                                  & Geometric DDPC        & 0.0779 [0.0648, 0.0866] & 2.381                & 6/6              \\
                                  & Direct DeePC          & 0.0731 [0.0594, 0.0771] & 1.977                & 6/6              \\
        \bottomrule
    \end{tabular}
    \caption{Cyclic tracking at \(\sigma_y=0.001\,\mathrm{rad}\). Errors are in radians and torque RMS in \(\mathrm{N\,m}\); \(n_{\mrm c}/6\) gives completed/attempted trials.}
    \label{tab:simulation:tracking}
\end{table}

With exact mechanics, double-integrator MPC and geometric DDPC give similar accuracy and effort, both outperforming direct DeePC in tracking. With parameter errors, geometric DDPC improves on double-integrator MPC in every paired trial at greater effort. It also has lower median error than direct DeePC on the base and shifted cycles, while direct DeePC has lower error and effort on the faster cycle. All tracking trials complete with inactive actuator limits.

\section{Discussion and Conclusions}
\label{sec:conclusions}
\label{sec:simulation:discussion}

This study examined geometric robust DDPC for a feedback-linearised planar 2R arm. The results identify a useful role for data when the mechanical model captures the main structure but leaves systematic prediction errors. Learning the resulting virtual-input--position response improves prediction and tracking under parameter errors. With exact mechanical parameters, the known double integrator already provides an effective controller. Performance relative to direct regularised DeePC depends on the reference, with different trade-offs between accuracy and torque effort.

The predictive settings also remained effective across the tested noise levels and arm parameters. This transfer used newly collected data and the corresponding mechanical-model and observer updates. Higher measurement noise increased tracking error and actuator activity, although every geometric-DDPC trial completed.

\clearpage
\bibliography{refs}

\subsection*{Generative AI Use}
Generative AI tools assisted with literature review, preparation of TikZ illustrations, simulation-code optimisation, and auditing the mathematical exposition/results.

\clearpage

\appendix
\section{Mechanical Model, State Estimation and Torque Application}
\label{app:estimation}

\paragraph{Inertia and velocity coupling.}
For the coordinates and link parameters in \Cref{sec:manipulator}, the inertia and velocity-coupling terms are
\begin{equation}
    M(q)=\begin{bmatrix}\alpha+2\beta\cos q_2&\delta+\beta\cos q_2\\\delta+\beta\cos q_2&\delta\end{bmatrix},\qquad
    \eta(q,v)=\beta\sin q_2\begin{bmatrix}-2v_1v_2-v_2^2\\v_1^2\end{bmatrix}.
    \label{eq:manipulator:inertia-coupling}
\end{equation}
Here \(\alpha=\mathcal I_1+\mathcal I_2+m_1r_1^2+m_2(l_1^2+r_2^2)\), \(\beta=m_2l_1r_2\), and \(\delta=\mathcal I_2+m_2r_2^2\).

The estimator uses the double-integrator matrices \(A,B,C\) in~\eqref{eq:manipulator:virtual-lti} and measured joint positions.

\paragraph{Position-based state estimation.}
A Kalman filter on the ideal double-integrator model estimates position and velocity~\citep{welch2000introduction}. Let \(\wh x_k^-\) and \(\wh x_k^+\) be the prior and corrected estimates, with covariances \(\Sigma_k^-\) and \(\Sigma_k^+\). The current measurement gives
\begin{equation}
    \begin{aligned}
        K_k        & =\Sigma_k^-C^\top(C\Sigma_k^-C^\top+\Sigma_y)^{-1},\qquad
        \wh x_k^+=\wh x_k^-+K_k(\wt y_k-C\wh x_k^-),                           \\
        \Sigma_k^+ & =(I_4-K_kC)\Sigma_k^-(I_4-K_kC)^\top+K_k\Sigma_yK_k^\top.
    \end{aligned}
    \label{eq:manipulator:observer-correction}
\end{equation}
After selecting \(\mu_k\), propagate \(\wh x_{k+1}^-=A\wh x_k^++B\mu_k\) and \(\Sigma_{k+1}^-=A\Sigma_k^+A^\top+Q^{\mrm{KF}}\), where \(Q^{\mrm{KF}}=B\Sigma_aB^\top\) and \(\Sigma_a\) is a chosen acceleration-error covariance. The covariance models uncertainty in the observer prediction; the plant has no added process noise.

Between measurements, inverse dynamics uses \(\wh x(t_k+s)=A(s)\wh x_k^++B(s)\mu_k\) for \(0\leq s<T_{\mrm s}\), with \(A(s),B(s)\) obtained by replacing \(T_{\mrm s}\) in~\eqref{eq:manipulator:virtual-lti} by \(s\). The initial estimate uses the reset position and zero velocity. Commanded virtual inputs drive the filter during collection, startup and tracking.

\paragraph{Torque application.}
Let \(T_\tau\) divide \(T_{\mrm s}\). At \(t_{k,m}=t_k+mT_\tau\), for \(m=0,\ldots,T_{\mrm s}/T_\tau-1\), compute \(\tau^{\mrm{cmd}}_{k,m}=\tau^{\mrm{id}}(t_{k,m})\). Given the preceding applied torque \(\tau^{\mrm{prev}}\), positive magnitude bounds \(\tau_{\max}\), and slew bounds \(r_\tau\), apply
\begin{equation}
    \begin{aligned}
        \tau^{\mrm{sat}}_{k,m} & =\clip(\tau^{\mrm{cmd}}_{k,m},-\tau_{\max},\tau_{\max}), \\
        \tau^{\mrm{act}}_{k,m} & =\tau^{\mrm{prev}}+
        \clip(\tau^{\mrm{sat}}_{k,m}-\tau^{\mrm{prev}},-T_\tau r_\tau,T_\tau r_\tau).
    \end{aligned}
    \label{eq:manipulator:actuator}
\end{equation}
Here \(\clip(z,a,b)\) truncates each component of \(z\) to its interval \([a_j,b_j]\). The torque is held until the next update. Torque holding and actuator modification enter \(\Delta\tau^{\mrm{act}}\) in~\eqref{eq:manipulator:residual}.

\section{Data Collection and Reference Construction}
\label{app:data}

\paragraph{Data collection.}
A PD tracking controller with acceleration feedforward follows an auxiliary sinusoid \(q^{\mrm{col}}(t)\) and adds random virtual excitation:
\begin{equation}
    \mu_k^{\mrm d}=\ddot q^{\mrm{col}}(t_k)
    +K_p(q^{\mrm{col}}(t_k)-\wh q_k^+)
    +K_d(\dot q^{\mrm{col}}(t_k)-\wh v_k^+)+e_k^{\mrm d}.
    \label{eq:manipulator:collection}
\end{equation}
The diagonal gains \(K_p,K_d\succ0\) are fixed. The sinusoid varies between episodes, and \(e_k^{\mrm d}\) is independent, zero-mean Gaussian excitation. The complete command also depends on measured feedback.

For the feedback-linearised dataset, the recorded command is \(\mu_k^{\mrm d}\). For the direct dataset, it is \(\tau_k^{\mrm d}=\wh M(\wh q_k^+)\mu_k^{\mrm d}+\wh h(\wh q_k^+,\wh v_k^+)\). Both use the supplied model and estimated state, with the input-holding conventions of \Cref{sec:manipulator:data}.

Position is sampled before the new command acts. The recorded pair is \((u_k,\wt y_k)\), where \(u_k=\mu_k\) for the feedback-linearised system and \(u_k=\tau_k^{\mrm{cmd}}\) for direct DeePC. Inputs are stored before actuator modification. An episode of \(T_e\geq L\) samples supplies \(T_e-L+1\) columns, giving \(n_g=\sum_e(T_e-L+1)\) offline windows. All windows remain within their episodes.

\paragraph{Reference construction.}
\label{app:data:references}
The evaluation references are
\begin{equation}
    q_j^{\mrm{ref}}(t)=q_{\mrm c,j}+a_j\chi(t/T_{\mrm r})\sin(\omega t+\phi_j),\qquad j=1,2,
    \label{eq:manipulator:cyclic-reference}
\end{equation}
with centre \(q_{\mrm c}\), amplitudes \(a_j\), phases \(\phi_j\), angular frequency \(\omega\) and ramp duration \(T_{\mrm r}\). The ramp \(\chi(z)=10z^3-15z^4+6z^5\) for \(0\leq z\leq1\), extended by zero before and one after this interval, gives continuous position, velocity and acceleration.

The desired position samples are \(q_k^{\mrm{ref}}=q^{\mrm{ref}}(t_k)\). A virtual-input reference consistent with the ideal sampled dynamics is obtained from \(v_0^{\mrm{ref}}=\dot q^{\mrm{ref}}(0)\) and
\begin{equation}
    \mu_k^{\mrm{ref}}=\frac{2}{T_{\mrm s}^2}(q_{k+1}^{\mrm{ref}}-q_k^{\mrm{ref}}-T_{\mrm s}v_k^{\mrm{ref}}),\qquad
    v_{k+1}^{\mrm{ref}}=v_k^{\mrm{ref}}+T_{\mrm s}\mu_k^{\mrm{ref}}.
    \label{eq:manipulator:reference}
\end{equation}
This defines a piecewise-constant-acceleration interpolant through the desired positions. Its discrete velocity can differ from the analytic reference derivative. Torque and torque-rate demands of this interpolant are checked through the ramp and periodic motion.

\paragraph{Initial history and sample timing.}
Before predictive tracking, the arm is held near the initial reference for \(\max(T_{\mrm{ini}},12)\) sampling intervals, using the collection controller with a constant reference and zero added excitation. The final plant and observer states of this hold initialise the tracking run.

At update \(k\), the current measurement corrects the observer while the predictor uses completed pairs through \(k-1\). The selected command is applied immediately in simulated time. After the interval, \((u_k,\wt y_k)\) replaces the oldest history pair.

For the direct coefficient reference, extend the desired history before time zero by \(q_j^{\mrm{ref}}=q_0^{\mrm{ref}}\), \(v_j^{\mrm{ref}}=0\), and \(\tau_j^{\mrm{ref}}=\wh h(q_0^{\mrm{ref}},0)\) for \(j<0\). At nonnegative samples, use the torque reference in \Cref{sec:manipulator:comparators} with the sampled velocity and acceleration in~\eqref{eq:manipulator:reference}. The measured history enters the predictive consistency terms separately.

\section{Geometric Representation and Predictive Solves}
\label{app:geometric}

\subsection{Affine, Dimensionless Trajectory Coordinates}
\label{app:geometric:coordinates}

Let \(\wt H_w^{\mrm d}\in\R[4L\times n_g]\) be the concatenated Hankel matrix of measured pairs \(\col(\mu_k,\wt y_k)\), ordered by sample. Fix positive scales \(s_\mu,s_y\in\R[2]\) for virtual input and position, and set \(D_s=I_L\otimes\diag(s_{\mu,1},s_{\mu,2},s_{y,1},s_{y,2})\). Define the mean window \(c\), scaled centred matrix \(Z\), and fitted basis by
\begin{equation}
    \begin{aligned}
        c & =\frac{1}{n_g}\wt H_w^{\mrm d}\mbf1_{n_g},\qquad
        Z=D_s^{-1}(\wt H_w^{\mrm d}-c\mbf1_{n_g}^{\top}),    \\
        d & =2L+4,\qquad \wh V=\svd_d(Z),\qquad
        \wh{\mathcal S}=\im\wh V.
    \end{aligned}
    \label{eq:manipulator:subspace}
\end{equation}
Here \(\svd_d\) returns the leading \(d\) left singular vectors. Centring limits the data rank to at most \(n_g-1\), so the fit requires \(n_g>d\) and an adequately excited \(d\)th direction. This rank follows the ideal four-state model and approximates the implemented response.

Each candidate \(\mathcal S\) describes the physical affine set \(c+D_s\mathcal S\). The mean, scales and fitted subspace remain fixed during a run. Scaling sets both the relative input/output weights and the geometry used to compare subspaces.

Let \(\wt w_j=\col(\mu_j,\wt y_j)\) and \(w_j^{\mrm{ref}}=\col(\mu_j^{\mrm{ref}},q_j^{\mrm{ref}})\). Stack the measured past and desired future, then transform them into the fitting coordinates:
\begin{equation*}
    \bar b_k=\col(\wt{\mbf w}_{k-T_{\mrm{ini}}:k-1},\mbf w_{k:k+N-1}^{\mrm{ref}}),\qquad
    b_k=D_s^{-1}(\bar b_k-c).
\end{equation*}
For the returned trajectory parameter and adverse subspace, decode
\begin{equation}
    \mbf w_k^{\mrm a}=c+D_sP_{\mathcal S_k^{\mrm a}}\xi_k,\qquad
    \mu_k=S_\mu\mbf w_k^{\mrm a},\qquad
    S_\mu=\begin{bmatrix}0_{2\times4T_{\mrm{ini}}}&I_2&0_{2\times(4N-2)}\end{bmatrix}.
    \label{eq:manipulator:command}
\end{equation}
The selector extracts the first future virtual input. Physical future tracking weights are the inverse squared channel scales.

\subsection{Prediction for Prescribed Inputs}
\label{app:geometric:prediction}

For the prediction tests, write a fitted trajectory as \(w(\zeta)=c+D_sV\zeta\), with \(\zeta\in\R[d]\). Let \(S_u\) select all \(2L\) input entries and \(S_{y,\mrm p}\) the \(2T_{\mrm{ini}}\) past output entries. Given the complete prescribed input stack \(\mbf u^{\mrm{test}}\) and past measured positions \(\wt{\mbf y}^{\mrm{test,p}}\), compute
\begin{equation}
    \begin{aligned}
        \min_{\zeta\in\R[d]}\quad & \norm{S_{y,\mrm p}w(\zeta)-\wt{\mbf y}^{\mrm{test,p}}}_{Q_{\mrm p}}^2 \\
        \text{subject to}\quad    & S_uw(\zeta)=\mbf u^{\mrm{test}},\qquad
        Q_{\mrm p}=I_{T_{\mrm{ini}}}\otimes\diag(s_{y,1}^{-2},s_{y,2}^{-2}).
    \end{aligned}
    \label{eq:simulation:prediction-fit}
\end{equation}
The future output entries are the forecast. For the known double integrator, use its exact behaviour and zero affine mean in the same solve. Future positions enter only the error calculation.

\subsection{Sine Chart and Scenario-Exchange Optimisation}
\label{app:geometric:solver}

The scenario-exchange method alternates a convex master solve with an adverse-subspace search. The master minimises the largest cost over its current subspaces. The adverse search adds a subspace with a larger cost at the returned \(\xi\). The master uses the bound \(\norm{\xi}_2\leq(\sqrt{b_k^\top Wb_k}+\norm{b_k}_2)/\rho\), which contains the sublevel set whose worst-case cost is no greater than the cost at \(\xi=0\).

If \(\wh V_\perp\) is an orthonormal basis of the complement of \(\wh{\mathcal S}\), the sine chart is
\begin{equation}
    V(X)=\wh V(I_d-X^\top X)^{1/2}+\wh V_\perp X,\qquad
    X\in\R[(4L-d)\times d],\quad \norm{X}_{\mrm F}\leq\rho.
    \label{eq:manipulator:sine-chart}
\end{equation}
For \(\rho<1\), the positive-definite square root exists and the resulting basis is orthonormal. Its chordal distance from the fitted subspace is \(d_{\mrm{ch}}(\im V(X),\wh{\mathcal S})=\norm{X}_{\mrm F}\). Multistart local searches operate on this closed chart ball. Its extremal value equals the open-ball supremum; the selected chart point is restricted to \(\norm{X}_{\mrm F}\leq(1-10^{-7})\rho\) before command extraction, keeping the executed projection strictly inside the open ball.

Warm starts and an invertible preconditioning of the master improve its numerical solution without changing the objective. Let \(J_k^{\mrm U}\) be the numerical upper estimate of the worst-case cost at the returned parameter, \(J_k^{\mrm L}\) the master-based lower estimate of the optimum, and \(F_k^{\mrm a}\) the cost at the returned adverse projection. The recorded relative gaps are
\begin{equation}
    \delta_k^{\mrm{out}}=\frac{\max\{0,J_k^{\mrm U}-J_k^{\mrm L}\}}{1+|J_k^{\mrm U}|},\qquad
    \delta_k^{\mrm{adv}}=\frac{\max\{0,J_k^{\mrm U}-F_k^{\mrm a}\}}{1+|J_k^{\mrm U}|}.
    \label{eq:manipulator:numerical-gaps}
\end{equation}
These floating-point estimates assess the numerical solve. A finite candidate can be applied at the iteration limit with an open gap, and its termination status is retained in the results. \Cref{tab:appendix:solver} lists the numerical settings.

\begin{table}[htbp]
    \centering\small
    \begin{tabular}{@{}ll@{}}
        \toprule
        Numerical setting                    & Value                                  \\
        \midrule
        Exchange iteration limit             & 20                                     \\
        Adverse search                       & Three starts, up to 80 iterations each \\
        Master / spectral-search limits      & 500 / 80 iterations                    \\
        Relative outer / adverse gap targets & \(10^{-4}\) / \(10^{-4}\)              \\
        Inner / master tolerances            & \(10^{-6}\) / \(10^{-9}\)              \\
        Extracted chart radius               & At most \((1-10^{-7})\rho\)            \\
        \bottomrule
    \end{tabular}
    \caption{Numerical settings for geometric DDPC. Warm starts are enabled.}
    \label{tab:appendix:solver}
\end{table}

Across the 6540 geometric updates, 404 meet both gap targets and 6136 reach the exchange-iteration limit. The median relative outer gap is \(0.453\), with maximum \(0.765\); the median adverse-search gap is \(1.19\times10^{-7}\), with maximum \(0.491\). These bound separations leave the suboptimality of the executed decisions unresolved. Every executed subspace lies strictly inside the radius-\(0.01\) ball.

\subsection{Double-Integrator MPC}
\label{app:comparators:mpc}

Let \(\mathcal O_N=\col(C,CA,\ldots,CA^{N-1})\). Define \(\mathcal T_N\) by its \((i,j)\) block: \(CA^{i-j-1}B\) for \(0\leq j<i<N\), and zero otherwise. The predicted output is \(\mbf y=\mathcal O_N\wh x_k^-+\mathcal T_N\mbf\mu\). With \(\bar Q=I_N\otimes Q\) and \(\bar R=I_N\otimes R\), the minimiser satisfies
\begin{equation}
    (\mathcal T_N^\top\bar Q\mathcal T_N+\bar R)\mbf\mu_k
    =\mathcal T_N^\top\bar Q(\mbf q_k^{\mrm{ref}}-\mathcal O_N\wh x_k^-)
    +\bar R\mbf\mu_k^{\mrm{ref}}.
    \label{eq:appendix:mpc-condensed}
\end{equation}
The positive-definite coefficient matrix is factored once; the reference and observer prior update the right-hand side.

\subsection{Reference-Centred Direct DeePC}
\label{app:comparators:direct}

Let \(H=\wt H^{\mrm d}\) be the current direct data matrix, including online windows. The reference centre is \(g_k^{\mrm{ref}}=H^\dagger r_k\), with \(r_k\) defined in~\eqref{eq:preliminaries:coefficient-reference}.

For a wide matrix, take a thin QR factorisation \(H^\top=Q_cR_c\), with \(Q_c^\top Q_c=I\), and set \(H_c=R_c^\top\). Then \(H=H_cQ_c^\top\). Decompose \(g=Q_c\zeta+g_\perp\), where \(Q_c^\top g_\perp=0\). Since \(g_k^{\mrm{ref}}\) belongs to the row space of \(H\), it can be written as \(Q_c\zeta_k^{\mrm{ref}}\), with \(\zeta_k^{\mrm{ref}}=H_c^\dagger r_k\). Hence
\begin{equation}
    Hg=H_c\zeta,\qquad
    \norm{g-g_k^{\mrm{ref}}}_2^2
    =\norm{\zeta-\zeta_k^{\mrm{ref}}}_2^2+\norm{g_\perp}_2^2.
    \label{eq:appendix:qr-centre}
\end{equation}
The omitted component changes no prediction and is zero at the optimum. Thus the reduction is exact for the reference-centred squared penalty. When compression is unnecessary, use \(Q_c=I\) and \(H_c=H\). The pseudoinverse is evaluated by a singular-value decomposition with relative cutoff \(10^{-12}\).

Partition \(H_c=\col(U_c^{\mrm p},Y_c^{\mrm p},U_c^{\mrm f},Y_c^{\mrm f})\) and set \(\bar Q=I_N\otimes Q\), \(\bar R^\tau=I_N\otimes R^\tau\). Let \(\mbf\tau_k^{\mrm{ref}}=\col(\tau_k^{\mrm{ref}},\ldots,\tau_{k+N-1}^{\mrm{ref}})\). The condensed quadratic has coefficient matrix and linear term
\begin{equation}
    \begin{aligned}
        A_g & =(Y_c^{\mrm f})^\top\bar QY_c^{\mrm f}
        +(U_c^{\mrm f})^\top\bar R^\tau U_c^{\mrm f}
        +\lambda_y(Y_c^{\mrm p})^\top Y_c^{\mrm p}+\lambda_gI, \\
        h_k & =(Y_c^{\mrm f})^\top\bar Q\mbf q_k^{\mrm{ref}}
        +(U_c^{\mrm f})^\top\bar R^\tau\mbf\tau_k^{\mrm{ref}}
        +\lambda_y(Y_c^{\mrm p})^\top\wt{\mbf y}_k^{\mrm{ini}}+\lambda_g\zeta_k^{\mrm{ref}}.
    \end{aligned}
    \label{eq:appendix:direct-quadratic}
\end{equation}
The stationarity and equality equations are
\begin{equation}
    \begin{bmatrix}A_g&(U_c^{\mrm p})^\top\\U_c^{\mrm p}&0\end{bmatrix}
    \begin{bmatrix}\zeta_k\\\nu_k\end{bmatrix}
    =\begin{bmatrix}h_k\\\mbf u_k^{\mrm{ini}}\end{bmatrix},
    \label{eq:appendix:direct-kkt}
\end{equation}
where \(\nu_k\) is a scaled equality multiplier. Cholesky factors and a Schur complement solve this system. Scaled equality and stationarity residuals are checked before accepting the command. Data updates change the factors; reference and history updates change the linear term.

\section{Simulation Parameters and Evaluation}
\label{app:simulation}

\subsection{Fixed Settings and Data Collection}

All links are uniform rods with \(g_0=9.81\,\mathrm{m\,s^{-2}}\). The plant and model parameters are given in \Cref{tab:simulation:arms}. \Cref{tab:simulation:settings} lists the shared controller and simulation settings. The observer uses \(\Sigma_y=\sigma_y^2I_2\), \(\Sigma_a=\sigma_a^2I_2\), and diagonal initial covariance obtained from the listed standard deviations.

\begin{table}[ht!]
    \centering\small
    \renewcommand{\arraystretch}{1.08}
    \begin{tabular}{@{}>{\raggedright\arraybackslash}p{0.33\linewidth}>{\raggedright\arraybackslash}p{0.64\linewidth}@{}}
        \toprule
        Setting                            & Value                                                                                                                                            \\
        \midrule
        Predictive / torque-update periods & \(T_{\mathrm s}=0.1\,\mathrm s\), \(T_\tau=0.01\,\mathrm s\)                                                                                     \\
        History, horizon and rank          & \(T_{\mathrm{ini}}=4\), \(N=8\), \(L=12\); \(d=28\) in \(\R[48]\)                                                                                \\
        Offline data per method and trial  & Eight episodes of 60 samples; 392 windows                                                                                                        \\
        Collection gains and excitation    & \(K_p=12I_2\), \(K_d=7I_2\); excitation standard deviation \(0.7\,\mathrm{rad\,s^{-2}}\) per joint                                               \\
        Baseline position noise            & \(\sigma_y=0.001\,\mathrm{rad}\) per joint                                                                                                       \\
        Observer uncertainty               & \(\sigma_a=1\,\mathrm{rad\,s^{-2}}\); initial position and velocity standard deviations \(0.01\,\mathrm{rad}\) and \(0.1\,\mathrm{rad\,s^{-1}}\) \\
        Geometric settings                 & \(\gamma=300\), \(\rho=0.01\); position scale \(0.1\,\mathrm{rad}\) and virtual-input scale \(2\,\mathrm{rad\,s^{-2}}\) per joint                \\
        Direct DeePC settings              & \(Q=100I_2\), \(R^\tau=0.3I_2\), \(\lambda_y=3\times10^3\), \(\lambda_g=100\); centre \(g_k^{\mrm{ref}}\)                                        \\
        Torque magnitude / slew limits     & \((15,5)\,\mathrm{N\,m}\) / \((300,100)\,\mathrm{N\,m\,s^{-1}}\)                                                                                 \\
        Plant integration                  & Fourth-order Runge--Kutta, \(0.005\,\mathrm s\) steps                                                                                            \\
        \bottomrule
    \end{tabular}
    \caption{Shared simulation settings. The transfer study changes the specified arm and sensor-noise covariance while retaining predictive tuning and the collection procedure.}
    \label{tab:simulation:settings}
\end{table}

The auxiliary sinusoids have uniformly sampled centres in \(( -\pi/2\pm0.06,\,0.4\pm0.08)\,\mathrm{rad}\), amplitudes in \([0.25,0.45]\) and \([0.35,0.60]\,\mathrm{rad}\), periods in \([5,7.5]\,\mathrm s\), and phases in \([0,2\pi)\). Episodes start at zero velocity with an independent \(0.02\,\mathrm{rad}\) Gaussian position perturbation. Direct DeePC retains all offline columns and up to 392 completed online windows.

Direct-DeePC weights were selected on four separate Arm~A trials with the base reference and baseline noise, favouring completed runs with low cyclic error. The selected weights, torque reference and reference-centred penalty are then used in every evaluation condition. No condition-specific retuning is performed.

\subsection{References and Performance Measures}

The references in \Cref{tab:simulation:references} have phases \((0,\pi/3)\) and a two-second ramp. Evaluation follows a 1.2-second stabilising hold and lasts for the ramp and two cycles: 14 seconds for the base and shifted references, and 11 seconds for the faster reference. Each condition uses six independent data/noise pairs, with measurement noise paired across controllers.

For a completed run of \(K\) intervals, define \(e_k=q(t_k)-q_k^{\mrm{ref}}\) and the periodic indices \(\mathcal K_{\mrm c}=\{k:0\leq k<K,\ t_k\geq T_{\mrm r}\}\). The reported measures are
\begin{equation}
    E_{\mrm{cyc}}=\sqrt{\frac{1}{2|\mathcal K_{\mrm c}|}\sum_{k\in\mathcal K_{\mrm c}}\norm{e_k}_2^2},\qquad
    \tau_{\mrm{rms}}=\sqrt{\frac{1}{2KT_{\mrm s}}\int_0^{KT_{\mrm s}}\norm{\tau^{\mrm{act}}(t)}_2^2\,\mathrm dt}.
    \label{eq:simulation:metrics}
\end{equation}
The factor two averages over the joints. Torque effort includes the actual held torque after actuator modification. A trial stops when either position error exceeds \(2\,\mathrm{rad}\), either speed exceeds \(10\,\mathrm{rad\,s^{-1}}\), or a numerical failure occurs. The terminal state is checked as well. Incomplete runs enter completion counts but not full-duration error or effort statistics.

\subsection{Prediction Tests and Numerical Accuracy}

For each test window, the error is \(\norm{\mbf q^{\mrm{pred}}-\mbf q^{\mrm{true}}}_2/\sqrt{2N}\), where the vectors stack future positions. Each independently fitted predictor uses three separate 60-sample test episodes, giving 15 non-overlapping windows. Take the median over windows for each fit, then the median and range across six fits. Data are recollected for each noise/mechanics condition, and both predictors solve~\eqref{eq:simulation:prediction-fit} with identical inputs and past measurements.

The dimension \(d=28\) is retained throughout. Supporting fits with \(d=30\) and \(32\) gave larger prediction errors and poorer conditioning on the same test records. A numerical-integration check at half the step size changed saved-interval positions by at most \(1.42\times10^{-9}\,\mathrm{rad}\), with unchanged actuator paths. Optimisation settings and termination statistics are reported in \Cref{app:geometric:solver}.

\end{document}